\pdfoutput=1

\documentclass[11pt]{article}

\usepackage[]{EMNLP2023}

\usepackage{times}
\usepackage{latexsym}

\usepackage[T1]{fontenc}
\usepackage{times}
\usepackage{latexsym}

\usepackage[T1]{fontenc}
\usepackage[utf8]{inputenc}

\usepackage{microtype}
\usepackage{graphicx}
\usepackage{inconsolata}
\usepackage{CJKutf8}

\usepackage[utf8]{inputenc}

\usepackage{microtype}

\usepackage{inconsolata}

\title{Sudowoodo: a Chinese Lyric Imitation System with Source Lyrics}

\author{Yongzhu Chang\textsuperscript{1}, Rongsheng Zhang\textsuperscript{1}\thanks{ \quad Corresponding Author}, Lin Jiang\textsuperscript{2}, Qihang Chen \textsuperscript{2} \\ \textbf{Le Zhang} \textsuperscript{1}, \textbf{Jiashu Pu} \textsuperscript{1} \\
{\textsuperscript{1} \normalsize Fuxi AI Lab, NetEase Inc., Hangzhou, China } \\
{\textsuperscript{2} \normalsize Music AV Lab, NetEase Inc., Hangzhou, China} \\
\texttt{\normalsize \{changyongzhu, zhangrongsheng\}@corp.netease.com}
}

\begin{document}
\begin{CJK}{UTF8}{gbsn}

\maketitle
\begin{abstract}


Lyrics generation is a well-known application in natural language generation research, with several previous studies focusing on generating accurate lyrics using precise control such as keywords, rhymes, etc. However, lyrics imitation, which involves writing new lyrics by imitating the style and content of the source lyrics, remains a challenging task due to the lack of a parallel corpus. In this paper, we introduce \textbf{\textit{Sudowoodo}}, a Chinese lyrics imitation system that can generate new lyrics based on the text of source lyrics. To address the issue of lacking a parallel training corpus for lyrics imitation, we propose a novel framework to construct a parallel corpus based on a keyword-based lyrics model from source lyrics. Then the pairs \textit{(new lyrics, source lyrics)} are used to train the lyrics imitation model. During the inference process, we utilize a post-processing module to filter and rank the generated lyrics, selecting the highest-quality ones. We incorporated audio information and aligned the lyrics with the audio to form the songs as a bonus. The human evaluation results show that our framework can perform better lyric imitation.  Meanwhile, the \textit{Sudowoodo} system and demo video of the system is available at \href{https://Sudowoodo.apps-hp.danlu.netease.com/}{Sudowoodo} and \href{https://youtu.be/u5BBT_j1L5M}{https://youtu.be/u5BBT\_j1L5M}.

\end{abstract}

\section{Introduction}

AI creative assistants are artificial intelligence systems that can learn from large amounts of text data to understand human language and culture and use this knowledge to create content such as story generation \cite{Alabdulkarim2021AutomaticSG, Zhu2020ScriptWriterNS}, poetry writing \cite{Guo2019JiugeAH, Liu2019DeepPA, Yang2019GeneratingCC}, grammar and spelling checking \cite{Patil2021SpellingCA}, etc. In addition, AI creative assistants can also assist in songwriting \cite{Potash2015GhostWriterUA, Zhang2020YoulingAA, Shen2019ControllingSM} by learning from numerous songs, understanding human emotional expression, and creating music in a similar writing style to humans. Previous research \cite{Castro2018CombiningLL, Watanabe2018AML, Manjavacas2019GenerationOH, Fan2019AHA, Li2020RigidFC, Zhang2020YoulingAA, Zhang2022QiuNiuAC} has focused on generating lyrics based on specified keywords (e.g., \textit{Snow}), lyrics styles, themes, or user input passages, which generate new lyrics with limited control over the content. 
However, in actual music production, users sometimes adapt excellent songs by adding their own creativity while remaining the original lyrical structure, resulting in new lyrics. This requires stronger control over the source lyrics such as text content, emotion, and fine-grained writing styles.



\begin{figure*}[h]
    \centering
    \includegraphics[width=\linewidth, scale=1.00]{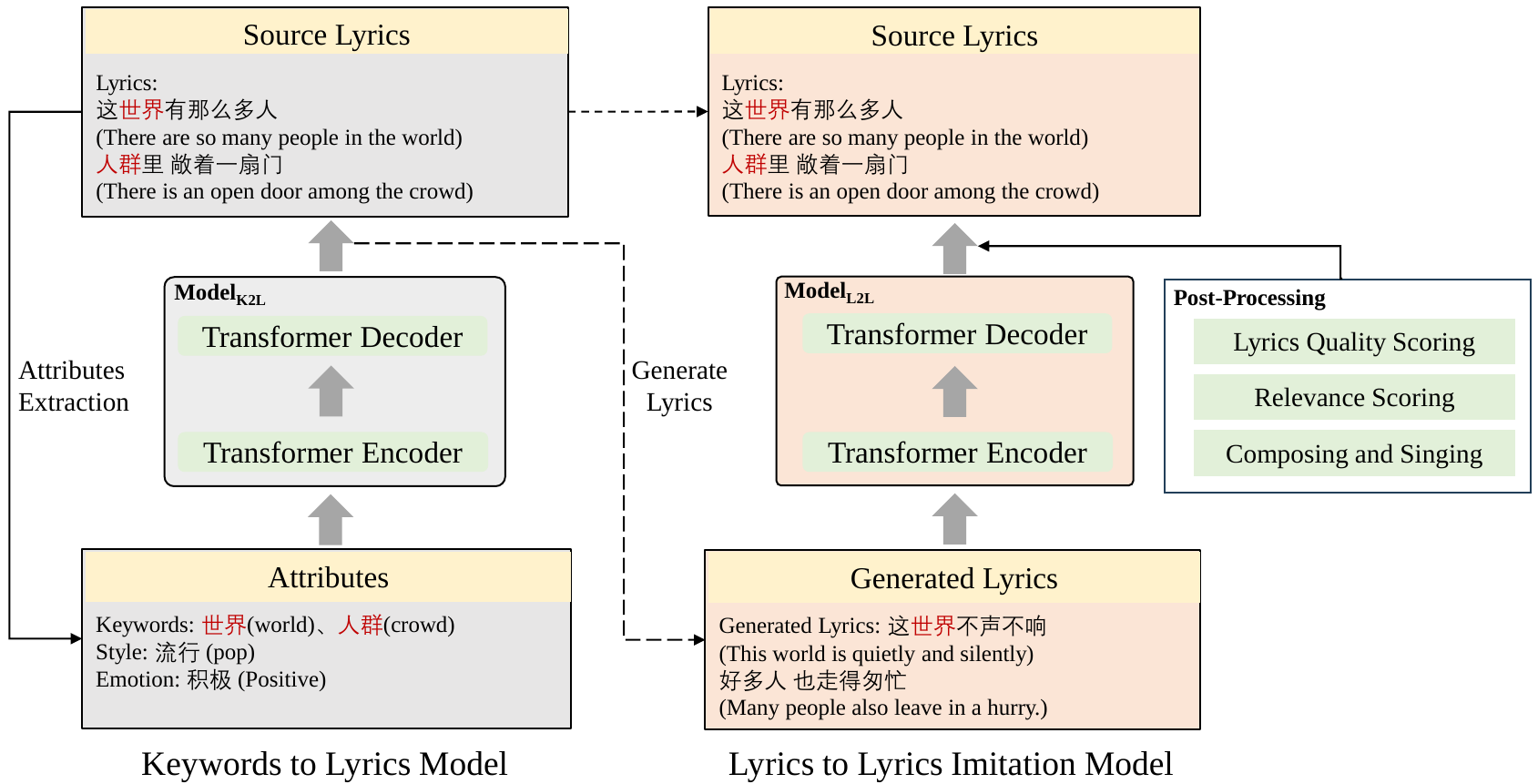}
    \caption{
       The framework of \textit{Sudowoodo} system proposes in this paper. \textbf{Model$_{K2L}$}  denotes a model for generating lyrics based on keywords, while \textbf{Model$_{L2L}$} represents the generation model from source lyrics to imitation lyrics.\textbf{Encoder} refers to the encoding portion of the Encoder-decoder architecture, while \textbf{Decoder} represents the decoding portion. \textbf{Post-processing} is mainly aimed at the imitation lyrics generated based on the Model$_{L2L}$.
    }
    \label{framte}
\end{figure*}

To address this issue, this paper demonstrates \textbf{\textit{Sudowoodo}} \footnote{https://en.wikipedia.org/wiki/Talk\%3ASudowoodo} (a Pokémon with the ability to imitate) a Chinese lyrics imitation generation system based on source lyrics. \textit{Sudowoodo} is typically based on the Encoder-Decoder framework, where the encoder encodes the text and attributes of the source lyrics, and the decoder generates the imitated lyrics. 
However, since we only have the source lyrics and not the target ones, the parallel corpus is lacking to train the imitation model. To solve the problem, we also propose a method for constructing aligned training samples, which generated the target lyrics from the extracted keywords of source lyrics using a keywords-based lyrics generation model. 

Specifically, we first collect the source lyrics corpus $D_{k}$ from the Internet \footnote{https://music.163.com/} and utilize the keyword extraction method described in Section~\ref{key} to extract keywords from source lyrics. And we train a keywords-based model, named Model$_{K2L}$, which can generate lyrics from given keywords. Then, we generate the target lyrics $D_{k'}$ using the Model$_{K2L}$. Finally, we train a lyrics imitation model with the aligned lyrics corpus ($D_{k'}$, $D_k$) based on the encoder-decoder framework. In addition, to improve the quality of generated lyrics and better showcase the results, we also employ post-processing modules including lyrics quality scoring and relevance scoring. Meanwhile, to provide a more intuitive understanding of the generated lyrics through imitation, we incorporate audio information (the vocals and melody of the source song) and align the lyrics with the audio to produce a complete song.

The main contributions of the \textit{Sudowoodo} system are summarized as follows:
\begin{itemize}
    \item We present a lyric imitation tool that generates new lyrics end-to-end based on source lyrics. Furthermore, we explore the addition of musical information to the generated lyrics in order to create songs. Sample songs can be heard at the songs of \href{https://Sudowoodo.apps-hp.danlu.netease.com/}{Sudowoodo}.
    \item We propose a novel framework for constructing a parallel lyrics corpus for imitation based on the keyword-based model. The results of the human evaluation show the efficacy of the imitation model trained on the basis of this parallel lyrics corpus.
    \item The \textit{Sudowoodo} system and demo video can be available at \href{https://Sudowoodo.apps-hp.danlu.netease.com/}{Sudowoodo} and \href{https://youtu.be/u5BBT_j1L5M}{https://youtu.be/u5BBT\_j1L5M}.
\end{itemize}

\section{Framework}

The \textit{Sudowoodo} system consists of two models and a post-processing module, as illustrated in Figure~\ref{framte}: \textbf{Model$_{K2L}$}, \textbf{Model$_{L2L}$}, and \textbf{Post-Processing}. These modules will be described in greater detail below.

\subsection{Data Preparation}
\label{key}

In this study, we obtain a dataset of $800k$ Chinese lyrics of various styles from the Internet, including pop, hip-hop, rap, etc. After filtering out lyrics less than 100 characters in length and removing duplicates, we are left with $600k$ unique lyrics. We denote the processed lyrics corpus as $D_k$. 

As depicted in the attribute extraction section of Figure~\ref{framte}, when conducting attributes extraction for the source lyrics, we extract not only the keywords of the source lyrics but other attributes such as style and emotion. To extract keywords from the source lyrics, we first segment the lyrics into multiple bars. We then apply KBERT \cite{Liu2019KBERTEL} based on distiluse-base-multilingual-cased-v1 \footnote{Multilingual knowledge distilled version of multilingual Universal Sentence Encoder. Supports 15 languages: Arabic, Chinese, Dutch, English, French, German, Italian, Korean, Polish, Portuguese, Russian, Spanish, and Turkish. https://www.sbert.net/docs/pretrained\_models.html} to extract a subset of the keywords from each bar. We extract 5 keywords for each bar. In addition, we rank the keywords according to their scores and select the top 10\% scoring keywords as the keywords for the whole song. In this process, we utilize the Jieba \footnote{https://github.com/fxsjy/jieba} as a word separation tool. For other information, we train a classifier model to acquire attributes such as emotion and style from source lyrics. Finally, we construct a parallel corpus dataset by extracting keywords, style, and emotion from the lyrics and aligning these attributes with source lyrics to form paired data ($D_A$, $D_K$), where $D_A$ represents the corpus composed of the extracted attributes of the corresponding source lyrics. The size of this dataset is $600k$.

\subsection{Models}

We first train a model, named Model$_{K2L}$, using the paired data ($D_A$, $D_K$) to generate lyrics based on keywords and their associated attributes such as emotion and style. Then, we acquire three new lyrics through Model$_{K2L}$ for each source lyric with random keywords extracted from the source lyric. The new lyrics are aligned with the source lyric and keywords to form paired data. All the lyrics generated by Model$_{K2L}$ are collected as $D_K'$. Consequently, we construct a parallel corpus dataset ($D_k'$, $D_k$) with a size of $1800k$. Meanwhile, during the training of Model$_{L2L}$, we encoder $D_k'$ and the write styles of $D_k$, while the decoding side targets $D_k$.


\textbf{Initialization:} To improve the model's performance and generate more fluent text, we initialize the model with a self-developed transformers-based pre-training model. Note that the structure of the pre-trained model is consistent with GPT-2 \footnote{https://openai.com/blog/gpt-2-1-5b-release/}, containing 210 million parameters with 16 layers, 1024 hidden dimensions, and 16 self-attention heads. The model is pre-trained on 30G of Chinese novels collected from the internet, using a vocabulary of 11400 words and a maximum sequence length of 512.

\textbf{Training:} Due to the lack of direct alignment corpus from lyrics to lyrics, we cannot train a seq2seq encoding and decoding model directly. Therefore, we propose a novel training strategy, as shown in Figure~\ref{framte}. The framework comprises two models for training. Firstly, a keyword-to-lyrics model, named Model$_{K2L}$, is used to generate aligned lyrics from source lyrics, with keywords and attributes such as style and emotion encoded into a latent semantic space and then decoded into source lyrics. The Model$_{K2L}$ utilizes an encoder-decoder architecture with the keywords, style, and emotion serving as encoder inputs and the source lyrics as decoder outputs, with training loss as shown in Equation~\ref{1-step}. Secondly, an end-to-end lyrics imitation model, called Model$_{L2L}$, is trained using the aligned corpus ($D_k'$, $D_k$) constructed from Model$_{K2L}$ and also utilizes the encoder-decoder architecture. The Model$_{L2L}$ encodes $D_k'$ and the attributes of the source lyrics into the encoder, with the source lyrics serving as the decoder output and training loss as shown in Equation~\ref{2-step}.


\begin{equation}
    L_{key2lyric}=-\sum_{D_k}log P(y_i|D(E(k_i, W_i)))
    \label{1-step}
\end{equation}

\begin{equation}
   L_{lyric2lyric}=-\sum_{(D_{k'} , D_k)}log P(y_i|D(E(x_i, k_i, W_i))) 
   \label{2-step}
\end{equation}

Where $E$ encodes lyrics, keywords, and writing styles into latent representation, and $D$ decodes the latent representation into lyrics. $k_i$ means the keywords and the $W_i$ represents the writing styles such as emotion and style in source lyrics. $x_i$ indicates the lyrics in $D_k'$. $D_k$ is the dataset of source lyrics, and $D_k'$ is lyrics generated from Model$_{K2L}$.

\textbf{Inference:} During inference, the input to Model$_{K2L}$ is controlled by keywords and writing style and is typically less than $512$ in length. In contrast, Model$_{L2L}$'s inputs include the source lyrics, which can easily exceed the length of $512$. The most intuitive approach is to truncate the inputs after incorporating the keywords and writing style. However, this approach would be easy to obscure the controlling elements such as writing style and keywords. To address this issue, when the lyrics exceed $512$ minus the length of the writing style and keywords, we truncate the last bar of the source lyrics to ensure that the input to the model does not exceed $512$. It is worth noting that the last bar of the source lyrics often repeats the previous content, so this truncation does not significantly impact the generated lyrics.

\textbf{Decoding Strategy:} We use a top-k sampling strategy with a sampling temperature of $0.8$ and a value of $k$ of $10$. Additionally, to prevent the model from easily generating duplicate words, we apply a sampling penalty technique proposed by \citet{Yadong2021KuiLeiXiAC}, which only penalizes the first $200$ words. In lyrics generation, although the model can learn the specific format, which is the number of lines and a number of words per line based on the source lyrics, we perform format control decoding to ensure that the generated lyrics have the same format as the source lyrics. To do this, we record the number of lines and words in the generated lyrics and adjust the \textit{[SEP]} and \textit{[EOS]} logits in each decoding step.





\subsection{Post-processing}

After the model training is finished, we can use the source lyrics, provided keywords, and writing styles to generate limitation lyrics with Model$_{L2L}$. We utilize the top-k sampling method at decoding to generate candidate lyrics. For each input, the model generates 10 samples. Then we re-rank the samples according to the following scores. 

\textbf{Lyrics Quality Scoring:} To filter high-quality lyrics, we train a classification model to determine whether a song lyric is a high-quality lyric and consider its confidence score as the Lyrics score, which is called $S_{Lyric}$, for re-rank. Inspired by QiuNiu \cite{Zhang2022QiuNiuAC}, we utilize popular and classic lyrics as positive samples, while lyrics with very few plays are negative samples. The experimental results indicate that the model gives a high confidence score when the lyrics contain beautiful sentences and rhetorical devices.

\textbf{Relevance Scoring:} In this paper, we introduce a method called $S_{relevance}$ to measure the semantic similarity between source lyrics and generated lyrics. To calculate $S_{relevance}$, we use the sentence transformer to obtain sentence vectors for both the source and generated lyrics, and then calculate the cosine similarity to rank the relevance. This method allows us to evaluate the quality of the generated lyrics in terms of their semantic similarity to the original lyrics.

Finally, we apply an anti-spam filter to the lyrics and use a combination of scores to sort them as shown in Equation~\ref{all_score}. We then select the top $3$ results as the final output. This post-process allows us to identify the most high-quality lyrics according to our criteria.

\begin{equation}
    \label{all_score}
    Score = w_1 * S_{Lyric} +  w_2 * S_{relevance}
\end{equation}

which the $w_1$ and $w_2$ denote the weights of the corresponding scores. In this paper, we set $w_1$ to $0.7$ and $w_2$ to $0.3$.



\begin{table*}[h]
    \centering
    \resizebox{0.8\textwidth}{!}{%
    \begin{tabular}{c|ccccc}
    \hline
                  & Theme (avg.) & Flu (avg.) & Logic (avg.) & Overall (avg.) & Best (\%) \\ \hline
    Model$_{K2L}$ & 4.168        & 4.103      & \textbf{3.480}        & 4.078          & 32.75     \\ \hline
    Model$_{L2L}$ & 4.250        & \textbf{4.160}      & 3.460        & \textbf{4.153} & \textbf{34.25}      \\
    w/o WS  & \textbf{4.275}        & 4.108      & 3.415        & 4.148          & 33        \\ \hline
    \end{tabular}%
    }
    \caption{Human evaluation results of Ablation. The scores in the table are the average scores of the three annotators. "Best" indicates that the model achieves Top-1 in the validation dataset for the same source lyric using three end-to-end lyrics imitation methods. \textit{Flu} means \textit{Fluency}, and \textit{Theme} is \textit{Thematic} in metrics. WS means the writing styles such as keywords, style, and emotion.}
    \label{result}
\end{table*}

\begin{figure}[h]
    \centering
    \includegraphics[width=\linewidth]{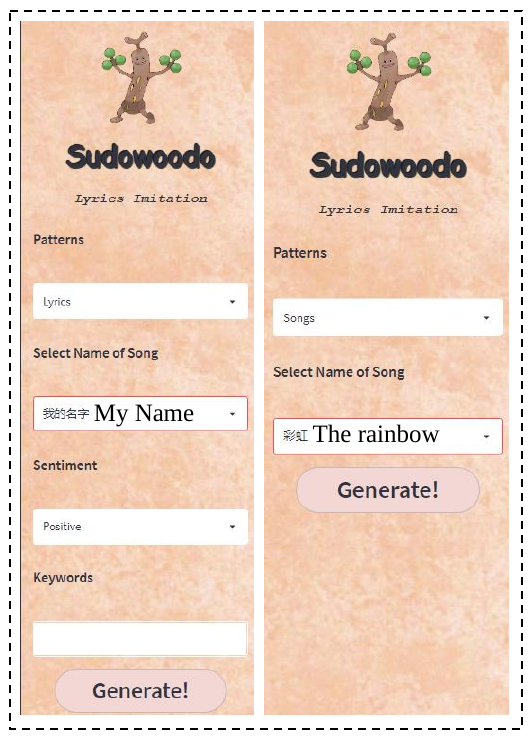}
    \caption{The interface of Sudowoodo.}
    \label{interface}
\end{figure}

\begin{figure*}[h]
    \centering
    \includegraphics[width=\linewidth, scale=1.00]{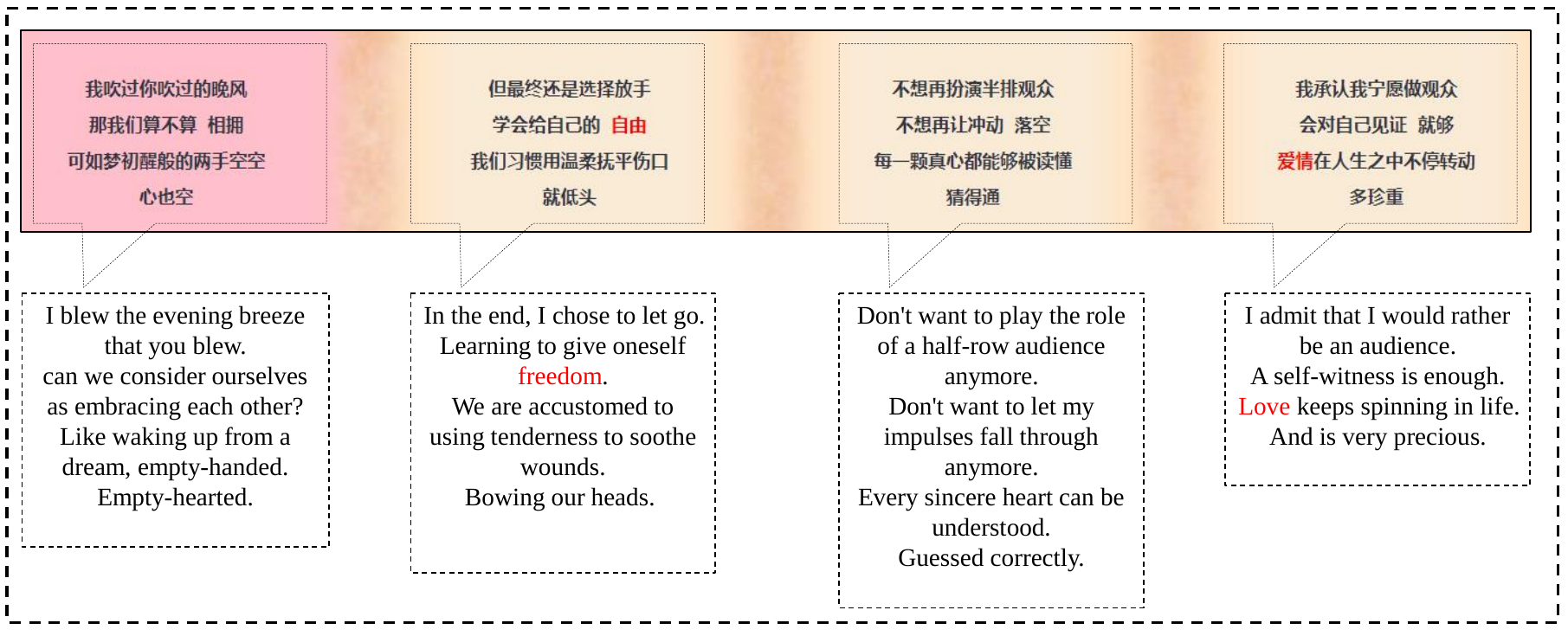}
    \caption{
    The instance of imitation lyrics in Lyrics mode. We enter the "爱情 (love)" and "自由 (freedom)" as keywords. As you can see from the picture, not all of the keywords entered are necessarily used. The Red color in Chinese and English indicates keywords.
    }
    \label{sample}
\end{figure*}

\begin{figure*}[h]
    \centering
    \includegraphics[width=\linewidth, scale=1.0]{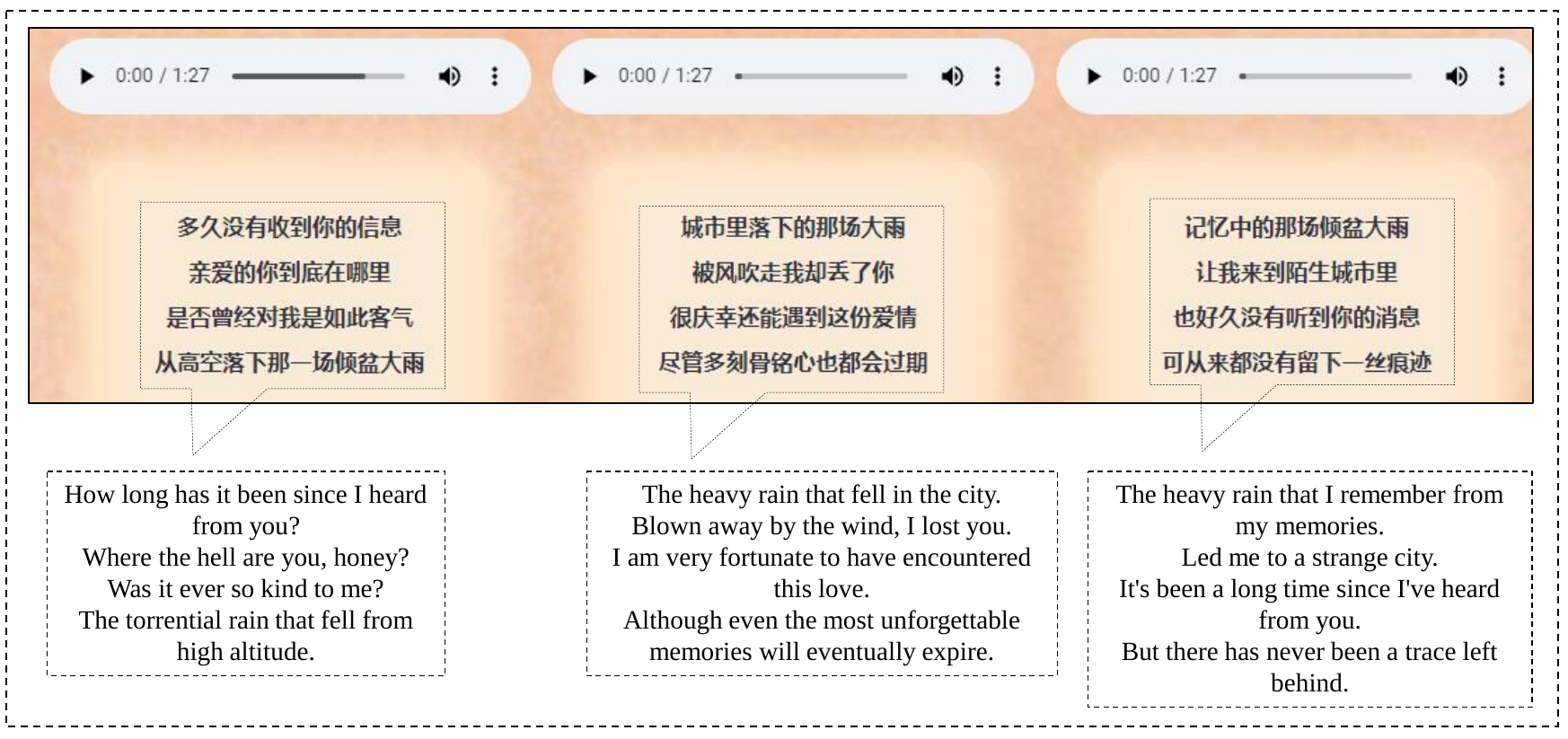}
    \caption{
    An example of Songs mode, with a player that plays the rendered song with imitation lyrics above the generated lyrics.
    }
    \label{songs}
\end{figure*}

\textbf{Composing and Singing:} In order to evaluate the quality of lyrics generated from the Model$_{L2L}$, we annotate popular songs by extracting various musical features, including melody, chord progression, key, structure, and phrasing, using both general music theory \footnote{https://www.ipr.edu/blogs/audio-production/what-are-the-basics-of-music-theory/} and more advanced analytical techniques. Based on these features, we then use intelligent composition \cite{Song2009AWF} techniques to generate melodies similar to those in the source style. Additionally, we use matching arrangement techniques, virtual vocal timbre selection, and mixing parameter adjustment to produce a fully synthesized song that includes accompaniment and singing. Finally, we incorporated audio information and aligned the lyrics with the audio to form the songs as shown in Figure~\ref{interface}. We can enjoy it in songs mode of \href{https://Sudowoodo.apps-hp.danlu.netease.com/}{Sudowoodo}!

\section{Results of the Experiment}

We conduct an ablation study to evaluate the framework proposed in this paper.

\textbf{Metrics:} We evaluate the generated lyrics from four perspectives: (1) \textit{Thematic:} The relevance of the imitation lyrics to the theme of the source lyrics, including love, friendship, family inspiration, etc. (2) \textit{Fluency:} It refers to the smoothness and naturalness of the language used in the lyrics. In evaluating the fluency of a song's lyrics, we consider factors such as the fluency of the words and the rhythmic structure of the sentences. (3) \textit{Logic:} It refers to the coherence and smoothness of scene transitions in the lyrics. To evaluate the logic of a song's lyrics, we consider whether consecutive sentences describe a single scene. If $m$ consecutive sentences describe a scene, we argue that those sentences are reasonable within logic. If $n$ consecutive groups of $m$ sentences are found to exist within $n$ different scenes, the lyrics are considered to have a high degree of smooth scene transitions overall. The number of scene jumps \footnote{Scene jumps occur when consecutive sentences describe different things or switch abruptly between different sensory perspectives, resulting in an unnatural or jarring transition.} can measure the logic of the song. (4) \textit{Overall:} The overall scoring of a song's lyrics.

\textbf{Results:} We sample $100$ lyrics from the source dataset and generate three imitation lyrics for each source lyric. We invite $3$ professional lyricists to score each of the 300 lyrics based on \textit{Thematic}, \textit{Fluency}, \textit{Logic}, and \textit{Overall}. The score ranges from 1-5, with 5 being the best and 1 being the worst. The results are shown in Table~\ref{result}, where all scores are averages for one song. We observe that the thematic and comprehensive scores of Model$_{L2L}$ exceeded those of Model$_{K2L}$. Additionally, we also verify the effect of the model, which uses only lyrics as input without keywords and writing style, and find that the addition of keywords improves the fluency of the generated lyrics. When the model is used to generate lyrics for the same lyrics using all three end-to-end methods, we observe that the method based on generated lyrics outperforms the keywords in 67.25\% of cases. It indicates that generated lyrics for training can improve the performance of a lyric imitation model.

\section{Demonstration}

This section demonstrates how the \textit{Sudowoodo} system works. 

The user interface for this demo is shown in Figure~\ref{interface}. As an imitation demo, it offers limited interaction with the user. The \textit{Sudowoodo} system operates in two modes: \textbf{\textit{Lyrics}} and \textbf{\textit{Songs}}. In \textbf{\textit{Lyrics}} mode, the user is required to select the source lyrics and the desired sentiment for the generated lyrics. Additionally, the user may provide keywords, which are typically space-separated phrases such as "自由 爱情 (\textit{freedom} \textit{love})" or, alternatively, left blank. When generating lyrics, the \textit{Sudowoodo} system takes into account the writing style of the selected source lyrics, including its theme, rhymes, and provided keywords, as well as the desired sentiment. The provided keywords are highlighted for easy identification. Note that not all provided keywords are necessarily used in the generated lyrics. In \textbf{\textit{Songs}} mode, the user can select the name of the source lyrics to hear the generated lyrics as a song. Due to technical limitations, the lyrics are rendered offline. In this paper, we apply three different AI singers to provide the sounds. Finally, the user can click "Generate!" to produce the output.



 
Next, we show some generated examples in Figure~\ref{sample}.

\textbf{Lyrics:} The leftmost column of the display lyrics represents the source lyrics selected by the user, while the three columns on the right show the generated imitation lyrics. If the user has entered keywords, these will be highlighted in red within the generated lyrics. This demo can generate smooth, high-quality lyrics in a format and writing style similar to the source lyrics for each generation.

\textbf{Songs:} Figure~\ref{songs} shows the results in Songs mode. As the real-time rendering of songs is a challenging task, we have performed offline rendering for this demo. A player is provided above the generated lyrics, which can be clicked on to hear the resulting song after rendering with the imitation lyrics. In the future, we aim to integrate real-time rendering of songs to create a true lyric imitation system that can take source lyrics and generate corresponding songs.
More experiences are available in \href{https://Sudowoodo.apps-hp.danlu.netease.com/}{Sudowoodo}.

\section{Conclusion}

In this paper, we describe \textit{Sudowoodo}, a Chinese lyric imitation system that supports two modes: Lyrics and Songs. In Lyrics mode, users can input keywords to generate imitated lyrics based on existing lyrics. In Songs mode, \textit{Sudowoodo} uses an unspecified technology to generate music that accompanies the imitated lyrics to create a complete song. To address the lack of a lyric-to-lyric alignment corpus, we propose a novel training framework structure to construct a parallel corpus for lyric imitation. Additionally, we apply Chinese pre-trained GPT-2 for initialization. To improve the quality of the generated lyrics, we employ a post-processing module to sort the generated results and select the highest quality ones. Finally, We audio-aligned some of the imitation lyrics to form songs! 



\bibliography{main.bib}
\bibliographystyle{acl_natbib}



\end{CJK}
\end{document}